\documentclass[11pt]{article}

\usepackage[utf8]{inputenc}
\usepackage[T1]{fontenc}
\usepackage[margin=1in]{geometry}
\usepackage{setspace}
\usepackage{graphicx}
\usepackage{float}
\usepackage{dblfloatfix}

\usepackage{subcaption}
\usepackage{multirow}
\usepackage{amsmath,amssymb,amsfonts}
\usepackage{mathrsfs}
\usepackage{xcolor}
\usepackage{textcomp}
\usepackage{booktabs}
\usepackage[numbers,sort&compress]{natbib}
\usepackage{siunitx}
\usepackage[colorlinks=true,allcolors=blue]{hyperref}
\usepackage{xr-hyper}
\usepackage[capitalize,english,nameinlink]{cleveref}
\usepackage[title]{appendix}

\begin{document}

\title{Reinforcement Learning Enables Autonomous Microrobot Navigation and Intervention in Simulated Blood Capillaries}

\author{
Jannik Drotleff$^{1,\dagger}$,
Samuel Tovey$^{1,\dagger}$,\\
Paul Hohenberger$^{1}$,
Christoph Lohrmann$^{1}$,\\
Julian Ho{\ss}bach$^{1}$,
Konstantin Nikolaou$^{1}$,
Christian Holm$^{1,*}$\\[0.7em]
$^{1}$Institute for Computational Physics, University of Stuttgart,\\
Allmandring 3, 70569 Stuttgart, Baden-W\"urttemberg, Germany\\
$^\dagger$These authors contributed equally to this work.\\
$^*$Correspondence: holm@icp.uni-stuttgart.de
}

\date{}

\maketitle
\begin{abstract}
Autonomous microrobots navigating biological vasculature could enable targeted drug delivery and thrombolysis, yet training control policies for realistic environments remains an open challenge.
Prior reinforcement learning (RL) studies of microrobotic navigation have been limited to idealized geometries that omit complex hydrodynamic flow fields, confined branching structures, and dense cellular obstacles found in vivo.
Here, we develop a physically grounded simulation of a blood capillary network, incorporating realistic hydrodynamic flow fields, explicit red blood cell dynamics, and anatomically derived branching geometry, and train deep RL agents to navigate it via chemotaxis.
We systematically map the physical limits of navigation across robot size and swimming speed, revealing a forbidden regime where Brownian motion and flow overcome propulsion.
Successful agents independently discover multiple universal strategy types, including run-and-rotate and energy-efficient search-and-sit policies, regardless of robot parameters.
Without retraining, these agents perform targeted blocking and unblocking of capillary flow, restoring throughput to healthy baseline levels.
These results establish RL as a viable framework for developing autonomous microrobotic intervention strategies in complex biological environments.
\end{abstract}
\clearpage
\twocolumn
\section{Introduction}\label{sec1}

The ability to precisely control robots on the micrometer scale would open a new era of medical applications, from microscale manufacturing~\citep{an25a} and porous media exploration~\citep{liao24a, lohrmann23a} to targeted drug delivery, minimally invasive surgery~\citep{lee23b}, and thrombolysis~\citep{lv25a}.
The clinical urgency is substantial: thrombosis-related conditions account for one in four deaths worldwide~\citep{raskob14a}.
Realizing this potential demands solutions to two coupled challenges: building microrobots that can propel themselves at low Reynolds numbers and developing control strategies that function in the complex, noisy environments these robots will encounter.

At the microscale, motion is governed by a low Reynolds number (\mbox{$Re$}), where viscous forces dominate inertial effects, and propulsion becomes fundamentally different from macroscopic locomotion.
Despite this constraint, hardware progress has been rapid.
Externally controlled magnetic microrobots have shown promising results, both as nanoparticle swarms~\citep{schuerle19a, yu18c, jiang23a} and individual colloids~\citep{landers25a}.
Bio-inspired flagella provide propulsion and orientation mechanisms~\citep{dreyfus05a, yamazaki04a, zhang09b}, while Janus particles offer asymmetric surface properties that enable control via magnetic fields and laser-driven self-thermophoresis~\citep{su23a, jiang10a}.
Most recently, a fully integrated robotic surface swimmer with on-board sensing, computation, memory, communication, and locomotion was realized at the \SI{200}{\micro\meter} scale~\citep{lassiter25a}, demonstrating that previously assumed engineering limitations can be overcome.
Yet even the most advanced designs offer only rudimentary actuation (typically translation and uncontrolled rotation) and remain confined to surface swimming, far from operating inside the body.
Further complicating control is Brownian motion, a stochastic noise arising from thermal interactions with the surrounding fluid, which constantly perturbs the robot's trajectory.

Reinforcement learning (RL) has become a powerful tool for discovering effective navigation strategies in these challenging environments~\citep{cai25a}.
Initial applications focused on optimizing navigation within fluidic disturbances: Colabrese et al.~\citep{colabrese17a} utilized Q-learning to enable gravitactic microswimmers to exploit vortex flows, outperforming passive strategies, while Nasiri and Liebchen~\citep{nasiri22a} employed the Advantage Actor-Critic algorithm to guide active particles through complex two-dimensional flow fields.
Moving beyond purely synthetic settings, Mui\~nos-Landin et al.~\citep{muinoslandin21a} demonstrated RL-based microrobotic control under real experimental conditions.
Other work has focused on emergent behaviors: Xiong et al.~\citep{xiong25c} and Tovey et al.~\citep{tovey23a-pre} showed that RL agents trained via actor-critic architectures can autonomously rediscover chemotactic behaviors, the oriented motion toward higher chemical concentrations, recovering the same run-and-tumble strategies used by bacteria~\citep{berg72a, watari10a, darnton07a}.
The mechanics underlying these learned policies have been characterized~\citep{tovey24a}, revealing how neural network responses to environmental inputs produce the observed motion patterns.
To facilitate such studies, Tovey et al.~\citep{tovey25c} introduced SwarmRL, an open-source platform that integrates RL with molecular dynamics simulations for active matter research.

However, a significant gap remains between these idealized simulations and the application-specific environments needed for microrobotic deployment.
Current studies focus on understanding control algorithms and swimming mechanisms in open domains, simple channels, or uniform flow fields, with little attention given to the realistic environments required to model actual deployment scenarios.
One such environment is that of a blood capillary network: a complex geometry of branching paths under the influence of flow fields and filled with passive blood cells.
Biological capillaries present severe challenges from both simulation and learning perspectives, as robots must navigate confined geometries where flow fields are further complicated by the dense presence of dynamic obstacles such as red blood cells (RBCs)~\citep{freund14a, marsden15a}.

Here, we bridge this gap by developing a physically grounded simulation of a blood capillary network to train and evaluate deep RL agents. 
Rather than aiming for a perfect biological representation, we employ a geometry that qualitatively captures essential properties—such as branching morphology and confined flow.
Our environment incorporates realistic branching geometry derived from anatomical illustration, steady-state hydrodynamic flow fields, explicit RBC dynamics, and Brownian fluctuations.
Within this framework, we map the physical boundaries of successful chemotactic navigation across robot size and speed, analyze the emergent control strategies, and demonstrate that trained agents can perform targeted blocking and unblocking of capillary flow without retraining---capabilities directly relevant to embolotherapy and thrombolysis.

\section{Results}\label{sec:Results}
\subsection*{Capillary environment and navigation challenge}
\begin{figure*}[!t]
    \centering
    \includegraphics[width=\textwidth]{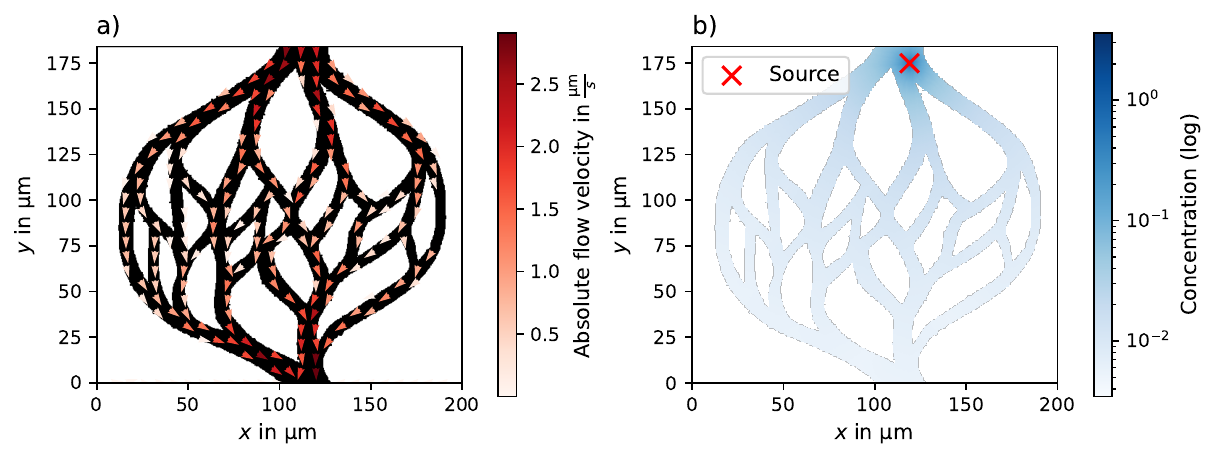}
    \caption{\textbf{Simulated capillary environment.} \textbf{a)} Capillary geometry and Lattice-Boltzmann flow field, derived from an anatomical capillary illustration~\cite{britannica26a}. \textbf{b)} Static concentration field for a chemical source at $\vec{r}_s = (\SI{119}{\micro \meter}, \SI{175}{\micro \meter})^T$, constrained by capillary walls.}
  \label{fig:system_overview}
\end{figure*}

To bridge the gap between idealized simulations and the complexity of biological environments, we established a realistic simulation environment modeled after human blood capillary structures.
We derived the simulation boundaries from an anatomical capillary illustration from the Encyclopedia Britannica~\cite{britannica26a}, mapping it into a domain of $\SI{200}{\micro \meter} \times \SI{184.13}{\micro \meter}$ based on physiological capillary diameters of \SI{5}{\micro\meter} to \SI{10}{\micro\meter}~\citep{betts13a}.
A steady-state flow field was computed using the Lattice-Boltzmann method, with the vessel geometry dictating local flow velocities and creating a training environment where peak velocities reach approximately \mbox{$3.5\times$} the mean flow (\hyperref[fig:system_overview]{\cref{fig:system_overview}a}). 

Central to our approach is chemotaxis, a directed motion along a chemical concentration gradient toward a source.
To represent the chemical signal, we applied a static concentration field that weakens as it spreads.
We constrained this field to the inside of the capillaries, ensuring that the gradient accurately traces the anatomical branching structure (\hyperref[fig:system_overview]{\cref{fig:system_overview}b}).
To maximize the navigation challenge for the RL agents, the chemical source was placed at \mbox{$\vec{r}_s = (\SI{119}{\micro \meter}, \SI{175}{\micro \meter})^T$}, in a region of comparatively high flow.
This steady-state approximation is justified by a P\'{e}clet number \mbox{$Pe < 1$} throughout the domain~\citep{bird02a}, ensuring that diffusion dominates over advective transport of the chemical signal (see \cref{sec:peclet} for the full P\'{e}clet analysis).

\subsection*{Physical limits of navigation}

\begin{figure*}[!t]
    \centering
    \includegraphics[width=\textwidth]{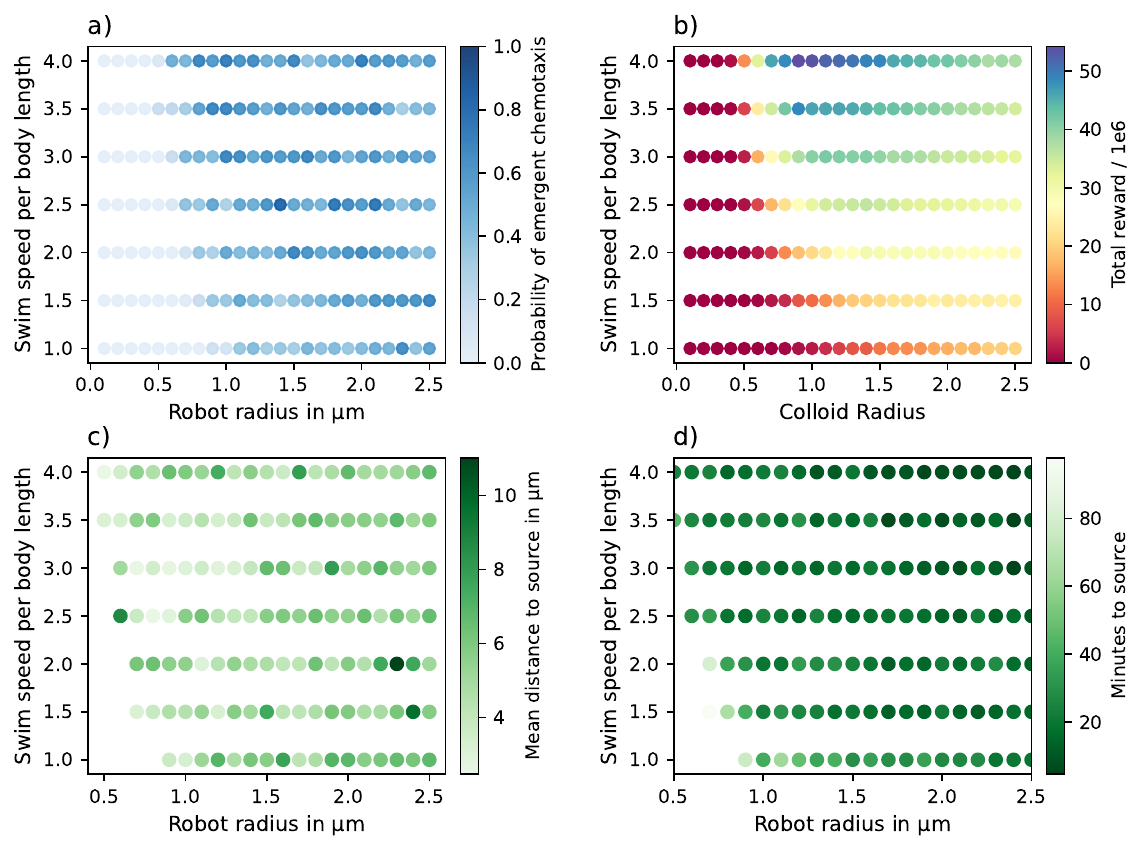}
    \caption{\textbf{Navigation performance across robot parameters.}
    \textbf{a)} Probability of successful chemotaxis (opacity scales with success rate); success requires \mbox{$\geq 8$} of 10 robots reaching within \SI{20}{\micro\meter} of the source.
    \textbf{b)} Cumulative reward across all 30 training runs per parameter combination.
    \textbf{c)} Mean equilibrium distance from the source.
    \textbf{d)} Mean time to reach the equilibrium distance.}
     \label{fig:training_results}
\end{figure*}

To map the physical boundaries of robotic navigation, we systematically varied the radius of the robot and the swimming speed, training 30 independent agents per combination of parameters with 10 robots each.
A successful agent was defined as one that steered at least eight of ten robots to within \SI{20}{\micro \meter} of the source during a two-hour deployment run.
The resulting phase diagram (\hyperref[fig:training_results]{\cref{fig:training_results}a}) reveals a distinctive forbidden regime for small, slow colloids, where Brownian motion and the opposing capillary flow overwhelm the robots' limited propulsive force.

Compared to an equivalent phase space analysis in a simpler environment with only Brownian motion~\citep{tovey24a}, this forbidden region shifts to larger and faster robots, as the complex boundary geometry, opposing flow, and RBC collisions form additional hindrances.
To confirm that this boundary reflects physics rather than algorithmic limitations, we transferred a single successfully trained model across the entire parameter space (\cref{sec:model-transfer}).
The persistent forbidden region under transfer confirms a fundamental physical boundary where environmental forces supersede any possible control strategy, and suggests that the RL agent finds a solution as soon as one is physically possible.

The highest success probability (\SI{80}{\percent}) was achieved by mid-sized robots (\mbox{$r = \SI{1.4}{\micro\meter}$}) at moderate speed (\mbox{$\SI{2.5}{\text{body lengths per second}}$}), indicating that maximum propulsion does not guarantee maximum success.
This reveals a clear size--speed trade-off.
Small robots (\mbox{$r < \SI{0.9}{\micro\meter}$}) are less affected by RBC collisions but are severely affected by Brownian fluctuations and local flow drag.
In contrast, large robots are resistant to noise and flow, but, given that the RBCs were modeled with \mbox{$r_\text{RBC} = \SI{1.5}{\micro\meter}$}, they cannot avoid collisions with these obstacles and lack maneuverability to navigate around them.

These physical trade-offs are also reflected in the learning dynamics.
The cumulative training reward (\hyperref[fig:training_results]{\cref{fig:training_results}b}), which indicates how efficiently the model learned the required policy, shows the same trade-off but with a peak for the fastest robots of the lower-mid-size.
This differs from the success rate results: while faster agents accumulate higher rewards by reaching the target quickly, they appear more susceptible to catastrophic failures in crowded flow, whereas larger mid-sized agents provide the physical stability necessary for consistent deployment success.

To characterize the actual performance of the swarm, we implemented successful strategies without further training and measured both the equilibrium distance and arrival time (\hyperref[fig:training_results]{\cref{fig:training_results}c,d}).
In contrast to findings in idealized bulk environments, equilibrium distances show no clear trend across the parameter space, ranging from \SI{2}{\micro\meter} to \SI{11}{\micro\meter}, comparable to values from our previous work despite substantially increased environmental complexity.
Although there is a slight tendency for smaller robots to get closer to the source, the physical constraints of capillary walls, RBC collisions, and opposing flow impose a floor on the equilibrium distance regardless of robot size, suggesting that the learned strategies remain effective at localized targeting even under realistic perturbations.
The time to arrival scales inversely with size and speed, but is roughly tenfold longer than in the simple Brownian case (\SIrange{10}{80}{\min} versus \SIrange{1}{10}{\min}), reflecting both the reduced net velocity from flow opposition and the tortuous paths imposed by the capillary geometry.

\subsection*{Emergent navigation strategies}

\begin{figure*}[!t]
    \centering
    \includegraphics[width=\textwidth]{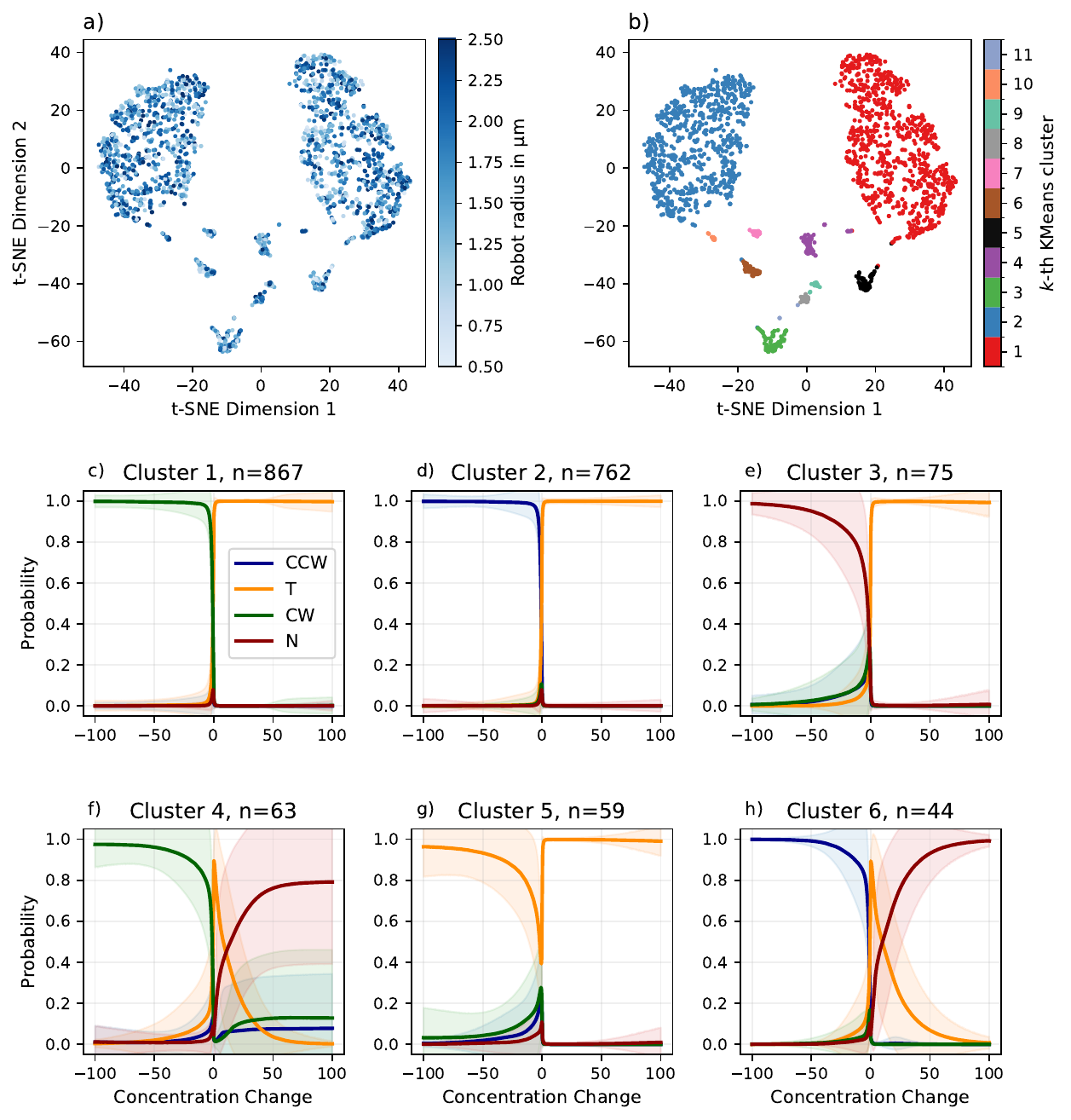}
  \caption{\textbf{Strategy analysis.}
  \textbf{a)} t-SNE embedding of learned policies colored by robot radius.
  \textbf{b)} Same embedding colored by $k$-means cluster assignment (\mbox{$k=11$}, selected by silhouette analysis~\citep{rousseeuw87a}).
  \textbf{c--h)} Mean action probability distributions (bold lines) and standard deviation (shaded) for the six largest clusters as a function of sensed concentration change.
  Positive values indicate motion toward the source; amplitude reflects proximity.
  Remaining clusters shown in \cref{sec:tsne-analysis}.}
  \label{fig:policies}
\end{figure*}

The above results demonstrate that RL can solve the challenge of navigation within complex biological environments.
To understand whether different robot configurations converge on similar control logic and what those strategies look like, we analyzed the learned actor networks directly rather than relying on trajectories alone, as different strategies do not necessarily produce distinguishable trajectories.

We mapped each agent's policy, defined as the action probability distribution across a range of chemical concentration changes, into a two-dimensional embedding using the t-distributed stochastic neighborhood embedding (t-SNE)~\citep{maaten08a} with perplexity 30 and PCA initialization, using the scikit-learn implementation~\citep{pedregosa11a} (\hyperref[fig:policies]{\cref*{fig:policies}a,b}).
Two prominent clusters emerge on the left and right of the embedding, with several smaller clusters below.
Critically, robots of all sizes appear in nearly every cluster (\hyperref[fig:policies]{\cref*{fig:policies}a}), and the same holds when coloring by swim speed (\cref{sec:strategy-analysis}), demonstrating that the emergent strategies are universal and independent of the robot's physical parameters.

To quantify the cluster structure, we applied $k$-means clustering~\citep{lloyd82a} with the silhouette metric~\citep{rousseeuw87a} to determine the optimal number of clusters, finding \mbox{$k_\text{max} = 11$} (\hyperref[fig:policies]{\cref*{fig:policies}b})
In the mean policy plots (\hyperref[fig:policies]{\cref*{fig:policies}c--h}), the sign of the concentration change denotes the direction of motion relative to the source (positive for approaching, negative for receding), while the amplitude indicates proximity: high-amplitude changes signify nearness to the source due to the steep reciprocal gradient, and low-amplitude changes indicate greater distance.

Analysis of individual strategy distributions reveals distinct navigational phenotypes.
The two dominant clusters (1 and 2) implement classic \textit{Run-and-Rotate} policies, characterized by clockwise or counterclockwise rotation for negative inputs and translation for positive inputs, with a transition around zero concentration change.
Cluster 3 employs \textit{Brownian Piloting}, an energy-efficient strategy where agents remain idle during large negative inputs and translate only when sensing small negative or large positive inputs, effectively conserving energy by relying on Brownian fluctuations for reorientation.

A distinct class of behavior emerges in Cluster 4 and its variants (Clusters 6--8, 10, and 11), which share a \textit{Search-and-Sit} phenotype specifically adapted to the capillary environment.
Here, agents translate with high probability only for small concentration changes, corresponding to the weaker gradients found far from the source.
As they approach the source, they transition to idling or rotating.
In effect, the agents swim up through the capillary system and, upon reaching the source, sit until the opposing flow drifts them away by a certain distance, at which point they repeat the approach.
This cycle emerges regardless of whether agents idle or rotate when being carried back, as these two actions are functionally equivalent in the presence of flow.

Finally, Cluster 5 (and the hybrid Cluster 9) follows a \textit{Gradient Gliding} strategy, maintaining a nearly continuous translation signal and using rotation only during weak gradient signals for course correction.
Although these types of strategy share heritage with policies documented in idealized environments~\citep{tovey24a}, they exhibit specific adaptations to capillary flow conditions, introducing a broader taxonomy of navigation tactics for confined, flow-dominated environments.
The remaining cluster strategies (7--11) are shown in \cref{sec:tsne-analysis}.

\subsection*{Functional intervention: blocking and unblocking capillary flow}

\begin{figure*}[!t]
    \centering
    \includegraphics[width=\textwidth]{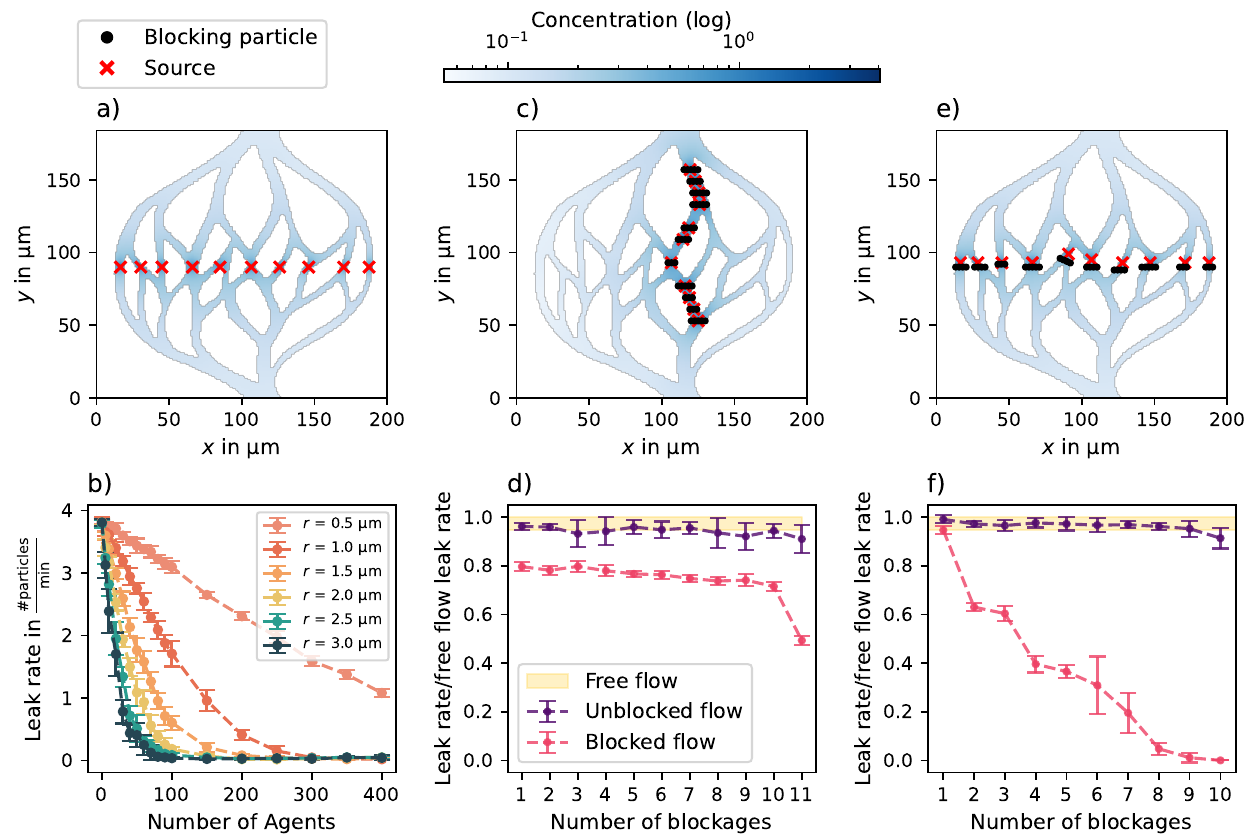}
    \caption{\textbf{Targeted capillary intervention.}
    \textbf{a)} Blocking task setup: chemical sources (red crosses) placed at \mbox{$y = \SI{90}{\micro\meter}$} in each vessel branch.
    \textbf{b)} Measured leak rate versus number of agents for varying robot radii; larger agents achieve full occlusion with fewer robots.
    \textbf{c,d)} Vertical unblocking setup and normalized leak rates showing near-complete flow restoration (green) relative to free flow (blue) and blocked baseline (red).
    \textbf{e,f)} Horizontal unblocking setup and corresponding leak rates.}
    \label{fig:Applications}
\end{figure*}

Although multiple universal strategy types emerged during training, we selected the Run-and-Rotate phenotype from Clusters 1 and 2 for functional task demonstrations, as it was the most prominent emergent strategy.
We now show how these behaviors can be leveraged for targeted medical interventions.
By strategically placing chemical sources, the microrobotic swarm can be commanded to perform functional tasks, specifically blocking and unblocking microvascular flow, without any retraining.
We note that it should be feasible to create the required local chemical concentration fields pharmacologically.
Because the learned policy relies solely on local gradient sensing, it can be applied independently of the source location.
These tasks mimic the requirements for embolotherapy, where blood flow is restricted to treat hemorrhages or starve tumors, and thrombolysis, where obstructions must be cleared.

For the \textit{blocking} task, we placed a chemical source in each tube of the capillary network at a height of \mbox{$y = \SI{90}{\micro\meter}$}, effectively commanding the robots to navigate there~(\hyperref[fig:Applications]{\cref*{fig:Applications}a}).
Varying both the radius of the robot and the size of the swarm, we measured the resulting rates of blood leak through the blockage (\hyperref[fig:Applications]{\cref*{fig:Applications}b}).
Blocking efficiency scales with both robot size and number: larger agents achieve total cessation of flow with fewer agents than smaller counterparts, owing to their greater cross-sectional area.
The leak rate decreases linearly with the number of agents until a critical density is reached, at which point the capillary becomes fully occluded.
These results demonstrate that the learned Run-and-Rotate policy is sufficiently robust to enable the collective formation of physical barriers, and that the system's flow resistance can be precisely tuned by modulating swarm composition.
This provides a versatile framework for autonomous embolotherapy.

For the inverse challenge of \textit{unblocking} (restoring flow to a pathologically obstructed network), we introduced multiple blockages strong enough to withstand the blood stream and deployed 20 robots per configuration.
We measured free flow (no blockages or robots), blocked flow (\mbox{$n_\text{blockages}$} present), and restored flow after robot deployment.
This was tested in two configurations: blockages added from top to bottom (\hyperref[fig:Applications]{\cref*{fig:Applications}c,d}) and from left to right (\hyperref[fig:Applications]{\cref*{fig:Applications}e,f}).
In both systems, the unblocked leak rates overlap with the free-flow baseline, as seen in the overlap between the blue and green regions in (\hyperref[fig:Applications]{\cref*{fig:Applications}d,f}).
This near-perfect flow restoration demonstrates complete recovery of capillary throughput by the microrobotic swarm, regardless of the initial number or arrangement of obstructions.
The Run-and-Rotate policy proves sufficient to clear even multi-point obstructions, effectively returning the capillary network to its healthy physiological state and demonstrating the capacity of RL to deliver autonomous, physically grounded solutions for in vivo microrobotic intervention.

\section{Discussion}\label{sec:discussion}

We have demonstrated that deep RL agents can learn robust universal navigation policies in a physically grounded capillary simulation and apply them, without retraining, to perform targeted vascular blocking and unblocking.
These results establish a concrete path from simulation-trained control to functional microrobotic intervention, contingent on corresponding advances in microrobot hardware.

The emergence of multiple distinct types of strategy has direct practical implications.
The energy-efficient Search-and-Sit and Brownian Piloting policies are particularly relevant given that on-board power at the microscale would be rapidly depleted in a flow environment.
Traditional power sources would be exhausted quickly; strategies that selectively idle and allow stochastic fluctuations or local flow patterns to assist in positioning offer a promising route to sustained operation.
That even the simplest Run-and-Rotate policy suffices for both blocking and unblocking tasks sets a performance floor and provides a baseline for future studies to investigate whether more specialized policies, such as Search-and-Sit, could offer higher efficiency under specific flow conditions.

For application demonstrations, we utilized only the Run-and-Rotate policy, as it emerged as the most prominent strategy.
Its success in solving medically relevant tasks suggests that even the most fundamental emergent behaviors are capable of addressing clinical problems.
Notably, because each agent relies solely on local gradient sensing, collective blocking emerges as a natural group behavior without requiring complex inter-robot communication---a significant practical advantage for swarm-based interventions.

Several simplifications in our simulation constrain the current level of realism, and addressing these represents a clear path for future work.

We model blood as a two-component fluid consisting of spherical RBCs and water-like plasma.
In nature, the characteristic RBC size is of order \SI{5}{\micro\meter}; they are not spherical but elliptical and deformable, capable of squeezing through the narrowest capillaries~\citep{phillips12a}.
To account for this, we used an effective diameter of \SI{3}{\micro \meter} to phenomenologically represent their high deformability.
We also used a rigid-wall approach, assuming that the shape of the vessel is not significantly affected by flow and particles~\citep{kannojiya21a}.

The F\r{a}hr\ae us-Lindqvist effect describes the migration of RBCs toward vessel centers, which leaves a cell-free layer near the walls~\citep{ascolese19a}. 
Although not explicitly modeled here, this phenomenon could actually benefit microrobot performance by allowing robots near the walls to avoid the highest concentration of RBCs and thereby reduce collisions.

A critical gap between our simulations and physiological conditions concerns flow velocity.
Our simulated flows are of order \SI{}{\micro\meter/s}, while physiological capillary blood flow ranges from \SI{0.47}{mm/s} to \SI{0.84}{mm/s}~\citep{bollinger74a}, with brain capillaries reaching \SI{0.5}{mm/s} to \SI{1.8}{mm/s}~\citep{hudetz97a}.
Since robots must swim at least as fast as the flow to overcome it, micron-sized robots would require swimming velocities of order \mbox{$10^3$} body lengths per second.
Our tests in this regime showed that the required angular velocities must increase by a similar factor, imposing a significant engineering challenge.
However, since physiological flow speeds are typically measured by tracking RBCs in the vessel center, the actual flow near the walls may be substantially slower.
Although most studies assume no-slip boundary conditions, recent in vivo measurements suggest wall-adjacent flow velocities of 30\% to 80\% of luminal velocity~\citep{jarolimova25a-pre}, which could relax the propulsion requirements for wall-following navigation strategies.

We also did not account for the pulsatile character of blood flow~\citep{koutsiaris03a}.
Future studies should implement non-static flow fields on a Lattice-Boltzmann foundation, though this would add significant computational cost (particularly when extending to three dimensions) and require addressing the shear-rate-dependent dynamic viscosity of blood~\citep{nader19a}.

At higher physiological flow speeds, the P\'{e}clet number increases and chemical signals will be advected downstream, producing smeared rather than static concentration fields.
The local-observable design of our agents makes them inherently well-suited for navigating such dynamic gradients, as the perceived concentration change that determines each action would naturally reflect the advected field.
However, it will be crucial to either model the concentration change noise realistically during in silico training or to develop precise sensors that offer accurate reception of the surrounding chemical concentration.

Importantly, the gradient-following logic learned by the RL agents is fundamentally gradient-agnostic: the same control architecture could operate on magnetic or acoustic gradients rather than chemical ones, potentially enabling external steering and control.
This flexibility substantially broadens the applicability of trained policies.

Despite the modeling simplifications discussed above, our results demonstrate that, given a sufficiently fast and sensitive agent, chemotaxis inside a blood flow environment is physically viable and functionally useful.
The simulation framework we have developed provides a foundation for increasingly realistic studies, while the emergent navigation strategies offer concrete design principles for future microrobotic systems.
As microrobot hardware continues to advance, the combination of RL-trained control policies with physically grounded simulation environments will be essential to bridge the gap between laboratory demonstrations and clinical deployment.

\section{Methods}\label{sec:methods}
\subsection*{Software}
The reinforcement learning experiments were performed using the open-source Python package SwarmRL~\citep{tovey25c}, with machine learning built on JAX~\citep{JAX} and Flax~\citep{flax}.
All molecular dynamics simulations used \textit{ESPResSo}~\citep{weik19a}, while the fluid flow was generated in \textit{waLBerla}~\citep{bauer21b}.

\subsection*{Simulation environment}
We generate a 2D boundary map by binarizing a capillary structure sketch from the Encyclopedia Britannica~\cite{britannica26a}.
The resulting image was cropped to $504\times464$ pixels, ensuring compatibility with the periodic boundary conditions that allow flow and particles to re-enter the capillary network.
Based on average capillary diameters of \SI{5}{\micro \meter} to \SI{10}{\micro \meter}~\citep{betts13a}, we set the system width to \SI{200}{\micro \meter}, which yields a height of \SI{184.13}{\micro \meter}, where each pixel represents a square of approximately \SI{0.40}{\micro \meter} side length.

Steady-state fluid dynamics was simulated using the Lattice-Boltzmann (LB) method in \textit{waLBerla}~\citep{bauer21b} with parameters listed in \cref{tab:lb-params}.
To adapt the 3D LB implementation to our 2D model, we extracted the central layer of the flow field and set all \mbox{$z$}-components to zero, creating a steady-state flow traversing the network from top to bottom.
The velocity fields were normalized to allow arbitrary scaling of the mean flow velocities for different agent configurations.
The peak velocities reach approximately \mbox{$3.5\times$} the mean flow velocity, creating a challenging dynamic environment where robots must constantly adapt to the varying drag forces.

\begin{table}[h]
\centering
\caption{\textbf{Parameters of the Lattice-Boltzmann simulation.}}
\begin{tabular}{r|c}
\toprule
\textbf{Parameter} & \textbf{Value} \\ \hline
Time step                  & 1.0                \\
Kinematic viscosity         & 0.2                \\
Density                     & 1.0                \\
Agrid                       & 1.0                \\
Tau                         & 1.0                \\
External force density           & $5 \times 10^{-6}$ \\
Inlet velocity              & 0.001              \\
Height                      & 20.0               \\
Length in $x$                     & 500.0              \\
Length in $y$                     & 462.738            \\
Flow direction               & {[}0, -1, 0{]}     \\
Number of steps                & 6960               \\
Resulting box dimensions                      & {[}504, 464, 22{]}
\\ \bottomrule
\end{tabular}
\label{tab:lb-params}
\end{table}
\subsection*{Chemical concentration field}
The chemical landscape was modeled as a reciprocal decay field, \mbox{$f(\vec{r}) = 1/d_s(\vec{r})$}, where $d_s$ represents the shortest path distance from the source to a given point $\vec{r}$, respecting wall boundaries.
This approach is justified by the assumption of a steady-state equilibrium in which the chemical has had sufficient time to diffuse throughout the capillary structure.
To verify this, we compute the P\'{e}clet number \mbox{$Pe = uL/D$}, where $u$ is the flow velocity, $L$ the system dimension, and $D$ the diffusion constant of the chemical attractant~\citep{bird02a}.
Assuming a glucose-like attractant with $D = \SI{7.7e-10}{m^2/s}$~\citep{liu97a}, we find \mbox{$5.181 \times 10^{-3} \leq Pe \leq 0.518$}, which confirms that diffusion dominates over advection for the concentration field.
To account for the complex geometry, we implemented a custom grid search algorithm that computes the shortest path around boundary obstructions, with the boundary map upscaled using \textit{scipy}~\citep{virtanen20b} for higher spatial resolution.

\subsection*{Particle dynamics}
Within this framework, blood is treated as a two-component fluid: plasma is modeled as a continuum, and red blood cells (RBCs) are treated as explicit particles.
Microrobot and RBC dynamics are governed by the overdamped Langevin equations:
\begin{align}
\label{eq:Langevin-translational}
    \dot{\vec{r}}_i &= \gamma_t^{-1}[F(t)\hat{e}_i(\theta_i)-\nabla V(\vec{r}_i, \{\vec{r}_j\})] \notag\\
    &\quad + \sqrt{2 k_B T\gamma_t^{-1}}\vec{R}^t_i(t) \\
\label{eq:Langevin-rotational}
    \dot{\theta}_i &= \gamma_r^{-1} m(t) \notag\\
    &\quad + \sqrt{2k_BT\gamma^{-1}_r}R^r_i(t),
\end{align}
where $\vec{r}_i$ and $\theta_i$ denote the position and orientation of particle $i$, $\gamma_t = 6\pi \mu r$, and $\gamma_r = 8 \pi \mu r^3$ are the translational and rotational Stokes friction coefficients, and $\mu$ is the dynamic viscosity.
We set \mbox{$\mu = \SI{0.89}{mPa \cdot s}$} near the \mbox{$\SI{1.2}{mPa \cdot s}$} value of blood plasma at \mbox{$T = \SI{37}{\degreeCelsius}$}~\citep{nader19a}; the minor discrepancy has negligible impact on the P\'{e}clet regimes of active and diffusive motion (see \cref{sec:peclet}).
$F$ and $m$ are the active force and torque, $\hat{e} = (\cos\theta, \sin\theta)^T$ is the orientation vector, $V$ is the interaction potential, $k_B$ is the Boltzmann constant and $\vec{R}_i^{(t,r)}$ are stochastic noise terms with $\langle R_i^{(t,r)}(t) R_j^{(t,r)}(t')\rangle = \delta_{ij}\delta(t-t')$.

Two-body interactions are modeled via the Weeks-Chandler-Andersen (WCA) potential~\citep{weeks71b}:
\begin{equation}\label{eq:WCA-definition}
V(r_{ij}) =
\begin{cases}
  4V_0\biggl[\Bigl(\tfrac{\sigma}{r_{ij}}\Bigr)^{12} - \Bigl(\tfrac{\sigma}{r_{ij}}\Bigr)^{6}\biggr] + V_0 \\
  \hfill \text{if } r_{ij} < r_c, \\[1ex]
  0 \hfill \text{else.}
\end{cases}
\end{equation}
with \mbox{$V_0 = k_B T$}, \mbox{$\sigma = (a_i + a_j) \cdot 2^{-1/6}$}, \mbox{$r_c = 2^{1/6}\sigma$}, and \mbox{$r_{ij} = ||\vec{r}_i - \vec{r}_j ||_2$}.
For the unblocking task, attractive bonds between blockage particles were implemented using a Lennard-Jones potential with \mbox{$\epsilon = \SI{76.94}{J}$}, \mbox{$\sigma_\text{block} = 2 r_\text{block} \cdot 2^{-1/6}$}, cutoff \mbox{$r_\text{cutoff} = 2.5\,\sigma_\text{block}$}, and \mbox{$r_\text{block} = \SI{1.2}{\micro \meter}$}, empirically tuned to withstand flow while remaining manipulable by agents.
All simulations used a time step of \mbox{$\tau = \SI{0.002}{s}$}.
The capillary geometry and steady-state LB flow field were integrated into the MD environment using \textit{ESPResSo}'s constraint features, with local flow velocity scaled to \SI{10}{\percent} of the agent's swimming velocity.

\subsection*{Reinforcement learning}
We employ an actor-critic architecture~\citep{barto83a, grondman12a} with Proximal Policy Optimization (PPO)~\citep{schulman17a-pre}, where the policy $\pi$ is represented by a neural network $\pi_\theta$.
The actor outputs an action probability distribution, while the critic estimates the value function $V(s_t)$.
The advantage is estimated using Generalized Advantage Estimation (GAE)~\citep{schulman18a-pre}, and the critic is optimized by minimizing the Huber loss~\citep{huber64a} between predicted values and target returns.
An entropy term encourages early exploration, yielding the total objective:
\begin{equation}
    \mathcal{L}_\text{total}= \mathcal{L}_\text{actor} - 0.5 \cdot \mathcal{L}_\text{critic} + \rho \cdot S,
\end{equation}
where \mbox{$S$} is the Shannon entropy and \mbox{$\rho = 0.02$}. The update is maximized via gradient ascent with \mbox{$\theta' = \theta + \eta \cdot \nabla_\theta \mathcal{L}$}.

We extend this to the multi-agent reinforcement learning (MARL) setting~\citep{gronauer22a}, where all agents share a single actor and critic network.
The experience collected by all agents is pooled for updates, with the collective objective:
\begin{equation}
    J_\text{MARL} = \frac{1}{N}\sum_{i=1}^{N_\text{agents}} J_i(\pi_\theta).
\end{equation}
The system follows a decentralized Markov decision process: each agent receives only local observations and individual rewards, sharing knowledge only through pooled training updates~\citep{oliehoek16a}.

The actor and critic networks each consist of two hidden layers with 128 nodes and ReLU activation, optimized with Adam~\citep{kingma17a} at the learning rate \mbox{$\eta = 0.002$}.
The discrete action space is
\begin{equation}
\mathcal{A}=
    \left\{
    \begin{aligned}
        \text{Translate:}\quad & v = n \cdot d,\ \omega = 0, \\
        \text{Rotate CCW:}\quad & v = 0,\ \omega = 10.472, \\
        \text{Rotate CW:}\quad & v = 0,\ \omega = -10.472, \\
        \text{Idle:}\quad & v = 0,\ \omega = 0.
    \end{aligned}
\right.
\end{equation}
Here, $v$ is measured in \si{\micro\meter/s} and $\omega$ in \si{rad/s}; $n$ sets the swimming speed relative to the diameter of the colloid $d$, motivated by the empirical scaling of the swimming speed with body size~\citep{murray92a}.
The rotation speed \mbox{$\omega = \SI{600}{\degree/s} = \SI{10.472}{rad/s}$} was inspired by \textit{Escherichia coli}~\citep{berg72a}.

The agent observable is the temporal change in perceived chemical concentration:
\begin{align}\label{eq:observable}
    o_i(t) = w \cdot \bigl[f\bigl(\vec{r}_i(t)\bigr) - f\bigl(\vec{r}_i(t-\Delta t)\bigr)\bigr],
\end{align}
with weight factor \mbox{$w = 100$}, concentration field $f$, and action interval \mbox{$\Delta t = \SI{0.1}{s}$}.
The reward is the clipped observable: \mbox{$r_i(t) = \max(0, o_i(t))$}, ensuring that robots are rewarded for approaching the source without explicit punishment for moving away.

For each size--speed combination, 30 independent training runs of 10{,}000 episodes were conducted with 10 robots each.
Each episode spans \SI{4}{s} of simulated time (40 policy applications at \mbox{$\Delta t = \SI{0.1}{s}$}), with the system reset after 1{,}000 uninterrupted episodes.
The robots were randomly initialized in the upper half of the capillary structure.
To prevent agents from becoming ``blind'' after being carried far from the source (where concentration changes approach zero), the system was reset when two agents drifted below \mbox{$y_\text{threshold} = -0.4 \times \SI{184}{\micro \meter}$}.
A custom checkpointer exported the model whenever the current reward exceeded the previous best, protecting against catastrophic forgetting in a noisy training environment.
The best-performing checkpoints were selected and evaluated in two-hour deployment simulations without further training.

For functional applications, we implemented a universal Run-and-Rotate policy.
It was trained with \mbox{$r = \SI{1.8}{\micro\meter}$} and \mbox{$\SI{2.5}{\text{body lengths per second}}$}.
In multi-source scenarios, concentration fields were computed as linear superpositions of individual fields.
During unblocking, target sources were deactivated after clearance and agents were turned off once all sources were cleared; final leak rates were measured only after agent deactivation to ensure they reflected the restored physiological state.

\subsection*{Computational resources}
Training and deployment were performed on University of Stuttgart compute resources, with each simulation utilizing one core of an AMD EPYC 9374F CPU node.
No GPU was required due to the relatively small network and system sizes.
Training required approximately six hours per model, and deployment simulations were completed in 90 minutes.

\section*{Data Availability}
The data generated and analyzed in this study will be made publicly available through the DaRUS repository upon publication.
\section*{Code Availability}
The analysis scripts used in this study will be made publicly available through the DaRUS repository upon publication. The SwarmRL framework is available as open-source software~\citep{tovey25c}.

\section*{Acknowledgments}
This study was funded by the Deutsche Forschungsgemeinschaft (DFG, German Research Foundation) through Compute Cluster grant no.\ 492175459.

\section*{Author Contributions}
S.T.\ conceived the project.
C.H.\ supervised the project.
S.T., J.D., and P.H.\ developed the training and deployment scripts.
P.H.\ developed the concentration field algorithm.
C.L.\ designed the capillary structure boundaries and performed the Lattice-Boltzmann simulations.
J.D.\ performed all RL agent training and deployment simulations, developed the application simulation scripts, and analyzed the data.
J.D.\ and S.T.\ wrote the manuscript and prepared the figures.
S.T., K.N., and J.H.\ provided project guidance and contributed to editing.
All authors reviewed the final manuscript.

\section*{Competing Interests}
The authors declare no competing interests.

\clearpage
\onecolumn
\begin{appendices}
\section{P\'{e}clet regimes}\label{sec:peclet}
\begin{figure}[H]
    \centering
    \includegraphics[width=0.9\linewidth]{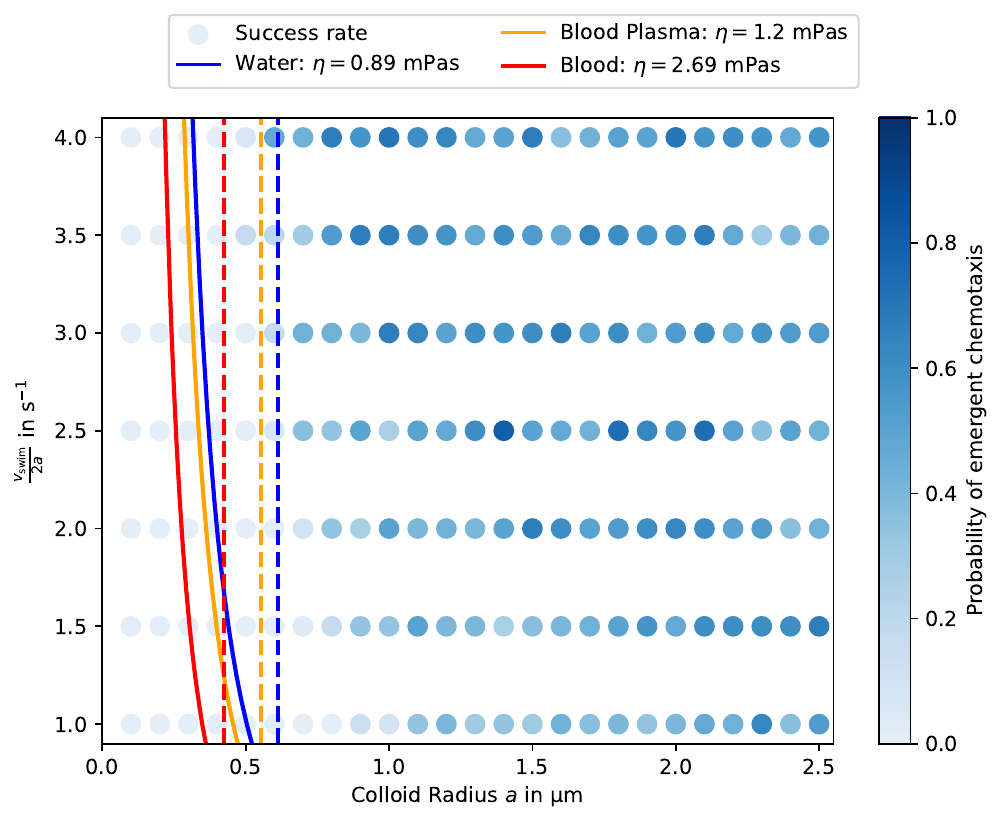}
    \caption{\textbf{P\'{e}clet analysis of the navigation phase space.}
    Translational and rotational P\'{e}clet boundaries (blue: water viscosity, orange: blood plasma viscosity) overlaid on the training success data.
    The forbidden regime broadly aligns with the diffusion-dominated P\'{e}clet regime, though the capillary flow field shifts the boundary relative to the pure Brownian case.
    The forbidden regime is reminiscent of the P\'{e}clet regime found in previous work~\citep{tovey24a}, where the P\'{e}clet number segments motion as either active-driven or diffusion-dominated.
    However, the capillary flow field prevents a pure diffusion description: the translational P\'{e}clet boundary does not precisely match the training success characteristics.
    The small difference between the P\'{e}clet lines computed with water viscosity ($\eta_\text{water} = \SI{0.89}{mPa \cdot s}$) and blood plasma viscosity ($\eta_\text{blood plasma} = \SI{1.2}{mPa \cdot s}$) confirms that this choice has negligible impact, producing at most a minor shift in the forbidden region boundary.}
    \label{fig:success_rate_peclet}
\end{figure}
\newpage

\section{Model transfer}\label{sec:model-transfer}
\begin{figure}[H]
    \centering
    \includegraphics[width=0.9\linewidth]{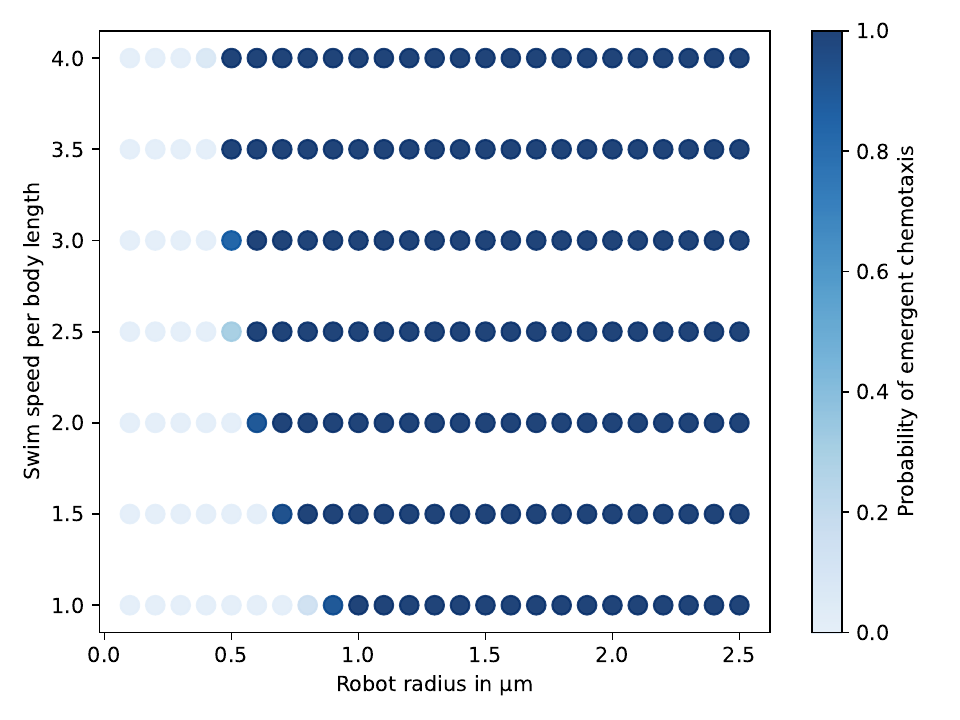}
    \caption{\textbf{Cross-parameter model transfer.} A single successfully trained policy was deployed across all robot sizes and speeds (30 simulations per combination). Color indicates the probability of successful chemotaxis ($\geq$8 of 10 agents within \SI{20}{\micro\meter} of the source). The persistent forbidden region confirms a fundamental physical boundary rather than an algorithmic limitation.}
    \label{fig:transfer_success_rate}
\end{figure}
\newpage

\section{Strategy analysis}\label{sec:strategy-analysis}
\begin{figure}[H]
    \centering
    \includegraphics[width=0.85\linewidth]{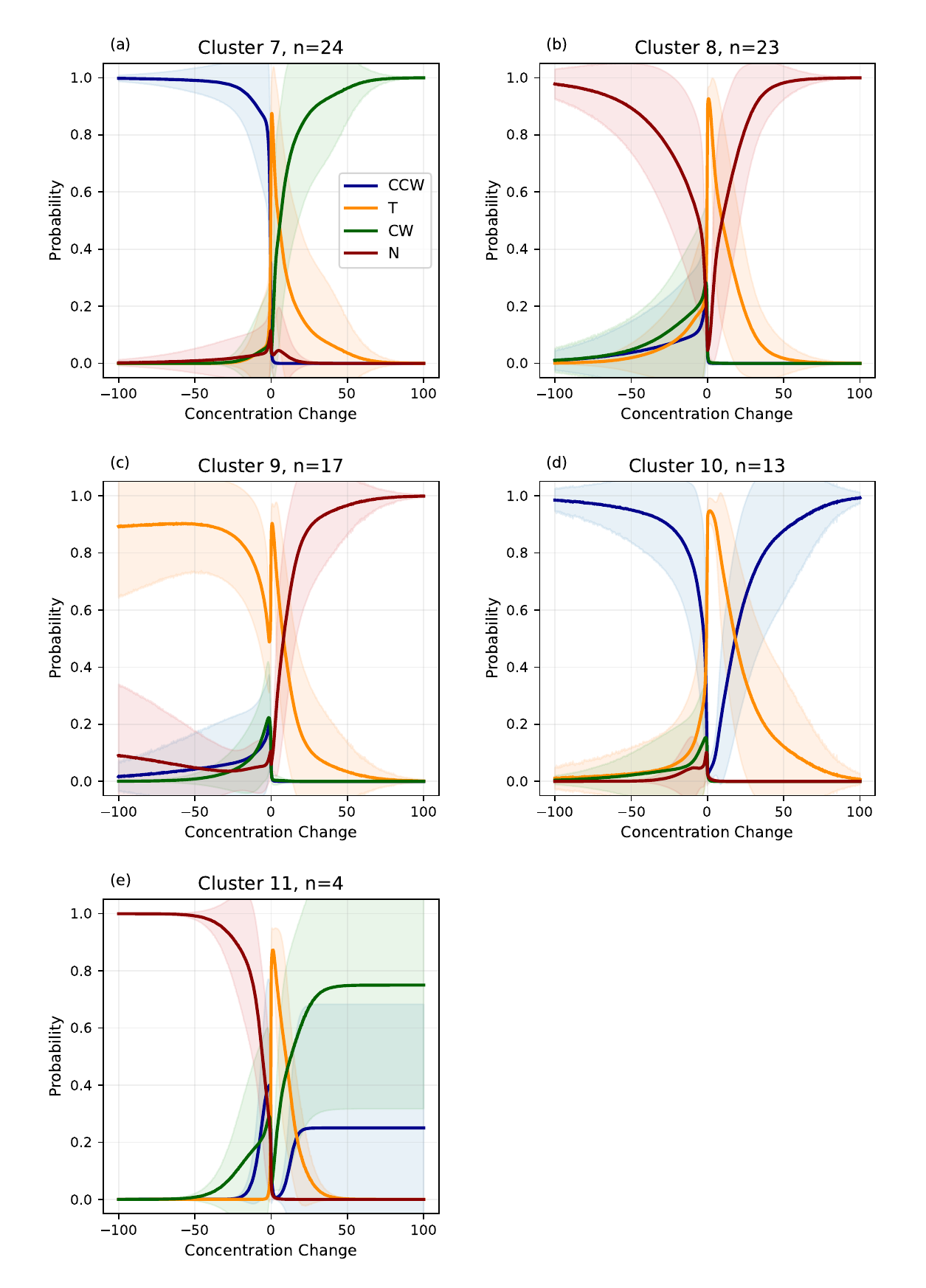}
    \caption{\textbf{Remaining strategy clusters.} Mean action probability distributions (bold lines) and standard deviation (shaded) for clusters 7--11, complementing the six largest clusters shown in the main text.}
\label{fig:appendix_remaining_strategies}
\end{figure}
\newpage

\section{t-SNE analysis for the swim multitudes}\label{sec:tsne-analysis}
\begin{figure}[H]
    \centering
    \includegraphics[width=0.9\linewidth]{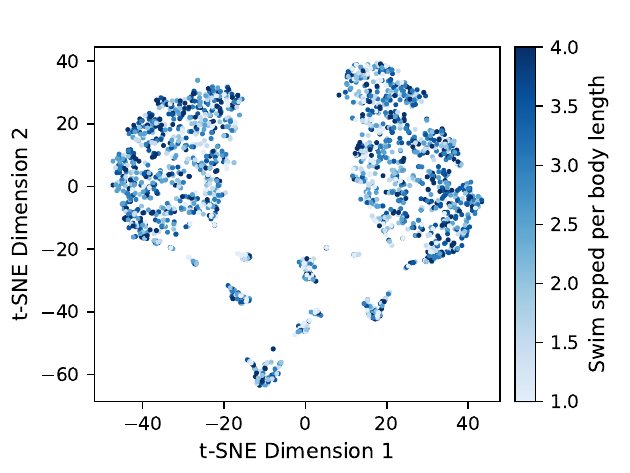}
    \caption{\textbf{t-SNE embedding colored by swimming speed.} The color indicates swim speed in body lengths per second. Robots of all speeds appear across clusters, confirming that emergent strategies are velocity-independent.}
\label{fig:tsne_swim_multitudes}
\end{figure}
\end{appendices}


\begin{thebibliography}{10}
\expandafter\ifx\csname url\endcsname\relax
  \def\url#1{\texttt{#1}}\fi
\expandafter\ifx\csname urlprefix\endcsname\relax\def\urlprefix{URL }\fi
\providecommand{\bibinfo}[2]{#2}
\providecommand{\eprint}[2][]{\url{#2}}

\bibitem{an25a}
\bibinfo{author}{An, Y.}, \bibinfo{author}{He, B.}, \bibinfo{author}{Ma, Z.},
  \bibinfo{author}{Guo, Y.} \& \bibinfo{author}{Yang, G.-Z.}
\newblock \bibinfo{title}{Microassembly: A review on fundamentals, applications
  and recent developments}.
\newblock \emph{\bibinfo{journal}{Engineering}} \textbf{\bibinfo{volume}{48}},
  \bibinfo{pages}{323--346} (\bibinfo{year}{2025}).

\bibitem{liao24a}
\bibinfo{author}{Liao, C.-T.} \emph{et~al.}
\newblock \bibinfo{title}{Propulsion of a three-sphere microrobot in a porous
  medium}.
\newblock \emph{\bibinfo{journal}{Physical Review E}}
  \textbf{\bibinfo{volume}{109}}, \bibinfo{pages}{065106}
  (\bibinfo{year}{2024}).

\bibitem{lohrmann23a}
\bibinfo{author}{Lohrmann, C.} \& \bibinfo{author}{Holm, C.}
\newblock \bibinfo{title}{Optimal motility strategies for self-propelled agents
  to explore porous media}.
\newblock \emph{\bibinfo{journal}{Physical Review E}}
  \textbf{\bibinfo{volume}{108}}, \bibinfo{pages}{054401}
  (\bibinfo{year}{2023}).

\bibitem{lee23b}
\bibinfo{author}{Lee, J.~G.}, \bibinfo{author}{Raj, R.~R.},
  \bibinfo{author}{Day, N.~B.} \& \bibinfo{author}{Shields, C. W.~I.}
\newblock \bibinfo{title}{Microrobots for biomedicine: {{Unsolved}} challenges
  and opportunities for translation}.
\newblock \emph{\bibinfo{journal}{ACS Nano}} \textbf{\bibinfo{volume}{17}},
  \bibinfo{pages}{14196--14204} (\bibinfo{year}{2023}).

\bibitem{lv25a}
\bibinfo{author}{Lv, S.}, \bibinfo{author}{Tang, L.~V.} \& \bibinfo{author}{Hu,
  Y.}
\newblock \bibinfo{title}{Application of nanotechnology and micro/nanorobots in
  thrombotic diseases}.
\newblock \emph{\bibinfo{journal}{EngMedicine}} \textbf{\bibinfo{volume}{2}},
  \bibinfo{pages}{100061} (\bibinfo{year}{2025}).

\bibitem{raskob14a}
\bibinfo{author}{Raskob, G.} \emph{et~al.}
\newblock \bibinfo{title}{Thrombosis: A major contributor to the global disease
  burden}.
\newblock \emph{\bibinfo{journal}{Journal of Thrombosis and Haemostasis}}
  \textbf{\bibinfo{volume}{12}}, \bibinfo{pages}{1580--1590}
  (\bibinfo{year}{2014}).

\bibitem{schuerle19a}
\bibinfo{author}{Schuerle, S.} \emph{et~al.}
\newblock \bibinfo{title}{Synthetic and living micropropellers for
  convection-enhanced nanoparticle transport}.
\newblock \emph{\bibinfo{journal}{Science Advances}}
  \textbf{\bibinfo{volume}{5}}, \bibinfo{pages}{eaav4803}
  (\bibinfo{year}{2019}).

\bibitem{yu18c}
\bibinfo{author}{Yu, J.}, \bibinfo{author}{Yang, L.} \& \bibinfo{author}{Zhang,
  L.}
\newblock \bibinfo{title}{Pattern generation and motion control of a
  vortex-like paramagnetic nanoparticle swarm}.
\newblock \emph{\bibinfo{journal}{The International Journal of Robotics
  Research}} \textbf{\bibinfo{volume}{37}}, \bibinfo{pages}{912--930}
  (\bibinfo{year}{2018}).

\bibitem{jiang23a}
\bibinfo{author}{Jiang, J.}, \bibinfo{author}{Yang, L.} \&
  \bibinfo{author}{Zhang, L.}
\newblock \bibinfo{title}{{{DQN-based}} on-line path planning method for
  automatic navigation of miniature robots}.
\newblock In \emph{\bibinfo{booktitle}{2023 {{IEEE International Conference}}
  on {{Robotics}} and {{Automation}} ({{ICRA}})}}, \bibinfo{pages}{5407--5413}
  (\bibinfo{year}{2023}).

\bibitem{landers25a}
\bibinfo{author}{Landers, F.~C.} \emph{et~al.}
\newblock \bibinfo{title}{Clinically ready magnetic microrobots for targeted
  therapies}.
\newblock \emph{\bibinfo{journal}{Science}} \textbf{\bibinfo{volume}{390}},
  \bibinfo{pages}{710--715} (\bibinfo{year}{2025}).

\bibitem{dreyfus05a}
\bibinfo{author}{Dreyfus, R.} \emph{et~al.}
\newblock \bibinfo{title}{Microscopic artificial swimmers}.
\newblock \emph{\bibinfo{journal}{Nature}} \textbf{\bibinfo{volume}{437}},
  \bibinfo{pages}{862--865} (\bibinfo{year}{2005}).

\bibitem{yamazaki04a}
\bibinfo{author}{Yamazaki, A.} \emph{et~al.}
\newblock \bibinfo{title}{Wireless micro swimming machine with magnetic thin
  film}.
\newblock \emph{\bibinfo{journal}{Journal of Magnetism and Magnetic Materials}}
  \textbf{\bibinfo{volume}{272--276}}, \bibinfo{pages}{E1741--E1742}
  (\bibinfo{year}{2004}).

\bibitem{zhang09b}
\bibinfo{author}{Zhang, L.} \emph{et~al.}
\newblock \bibinfo{title}{Characterizing the swimming properties of artificial
  bacterial flagella}.
\newblock \emph{\bibinfo{journal}{Nano Letters}} \textbf{\bibinfo{volume}{9}},
  \bibinfo{pages}{3663--3667} (\bibinfo{year}{2009}).

\bibitem{su23a}
\bibinfo{author}{Su, H.}, \bibinfo{author}{Li, S.}, \bibinfo{author}{Yang,
  G.-Z.} \& \bibinfo{author}{Qian, K.}
\newblock \bibinfo{title}{Janus micro/nanorobots in biomedical applications}.
\newblock \emph{\bibinfo{journal}{Advanced Healthcare Materials}}
  \textbf{\bibinfo{volume}{12}}, \bibinfo{pages}{2202391}
  (\bibinfo{year}{2023}).

\bibitem{jiang10a}
\bibinfo{author}{Jiang, H.-R.}, \bibinfo{author}{Yoshinaga, N.} \&
  \bibinfo{author}{Sano, M.}
\newblock \bibinfo{title}{Active motion of a {{Janus}} particle by
  self-thermophoresis in a defocused laser beam}.
\newblock \emph{\bibinfo{journal}{Physical Review Letters}}
  \textbf{\bibinfo{volume}{105}}, \bibinfo{pages}{268302}
  (\bibinfo{year}{2010}).

\bibitem{lassiter25a}
\bibinfo{author}{Lassiter, M.~M.} \emph{et~al.}
\newblock \bibinfo{title}{Microscopic robots that sense, think, act, and
  compute}.
\newblock \emph{\bibinfo{journal}{Science Robotics}}
  \textbf{\bibinfo{volume}{10}}, \bibinfo{pages}{eadu8009}
  (\bibinfo{year}{2025}).

\bibitem{cai25a}
\bibinfo{author}{Cai, W.}, \bibinfo{author}{Wang, G.}, \bibinfo{author}{Zhang,
  Y.}, \bibinfo{author}{Qu, X.} \& \bibinfo{author}{Huang, Z.}
\newblock \bibinfo{title}{Reinforcement learning for active matter}.
\newblock \emph{\bibinfo{journal}{Biophysics Reviews}}
  \textbf{\bibinfo{volume}{6}}, \bibinfo{pages}{031302} (\bibinfo{year}{2025}).

\bibitem{colabrese17a}
\bibinfo{author}{Colabrese, S.}, \bibinfo{author}{Gustavsson, K.},
  \bibinfo{author}{Celani, A.} \& \bibinfo{author}{Biferale, L.}
\newblock \bibinfo{title}{Flow navigation by smart microswimmers via
  reinforcement learning}.
\newblock \emph{\bibinfo{journal}{Physical Review Letters}}
  \textbf{\bibinfo{volume}{118}}, \bibinfo{pages}{158004}
  (\bibinfo{year}{2017}).

\bibitem{nasiri22a}
\bibinfo{author}{Nasiri, M.} \& \bibinfo{author}{Liebchen, B.}
\newblock \bibinfo{title}{Reinforcement learning of optimal active particle
  navigation}.
\newblock \emph{\bibinfo{journal}{New Journal of Physics}}
  \textbf{\bibinfo{volume}{24}}, \bibinfo{pages}{073042}
  (\bibinfo{year}{2022}).

\bibitem{muinoslandin21a}
\bibinfo{author}{{Mui{\~n}os-Landin}, S.}, \bibinfo{author}{Fischer, A.},
  \bibinfo{author}{Holubec, V.} \& \bibinfo{author}{Cichos, F.}
\newblock \bibinfo{title}{Reinforcement learning with artificial
  microswimmers}.
\newblock \emph{\bibinfo{journal}{Science Robotics}}
  \textbf{\bibinfo{volume}{6}}, \bibinfo{pages}{eabd9285}
  (\bibinfo{year}{2021}).

\bibitem{xiong25c}
\bibinfo{author}{Xiong, T.}, \bibinfo{author}{Liu, Z.}, \bibinfo{author}{Wang,
  Y.}, \bibinfo{author}{Ong, C.~J.} \& \bibinfo{author}{Zhu, L.}
\newblock \bibinfo{title}{Chemotactic navigation in robotic swimmers via
  reset-free hierarchical reinforcement learning}.
\newblock \emph{\bibinfo{journal}{Nature Communications}}
  \textbf{\bibinfo{volume}{16}}, \bibinfo{pages}{5441} (\bibinfo{year}{2025}).

\bibitem{tovey23a-pre}
\bibinfo{author}{Tovey, S.} \emph{et~al.}
\newblock \bibinfo{title}{Environmental effects on emergent strategy in
  micro-scale multi-agent reinforcement learning} (\bibinfo{year}{2023}).
\newblock \eprint{2307.00994}.

\bibitem{berg72a}
\bibinfo{author}{Berg, H.~C.} \& \bibinfo{author}{Brown, D.~A.}
\newblock \bibinfo{title}{Chemotaxis in escherichia coli analysed by
  three-dimensional tracking}.
\newblock \emph{\bibinfo{journal}{Nature}} \textbf{\bibinfo{volume}{239}},
  \bibinfo{pages}{500--504} (\bibinfo{year}{1972}).

\bibitem{watari10a}
\bibinfo{author}{Watari, N.} \& \bibinfo{author}{Larson, R.~G.}
\newblock \bibinfo{title}{The hydrodynamics of a run-and-tumble bacterium
  propelled by polymorphic helical flagella}.
\newblock \emph{\bibinfo{journal}{Biophysical journal}}
  \textbf{\bibinfo{volume}{98}}, \bibinfo{pages}{12--17}
  (\bibinfo{year}{2010}).

\bibitem{darnton07a}
\bibinfo{author}{Darnton, N.~C.}, \bibinfo{author}{Turner, L.},
  \bibinfo{author}{Rojevsky, S.} \& \bibinfo{author}{Berg, H.~C.}
\newblock \bibinfo{title}{On torque and tumbling in swimming escherichia coli}.
\newblock \emph{\bibinfo{journal}{Journal of Bacteriology}}
  \textbf{\bibinfo{volume}{189}}, \bibinfo{pages}{1756--1764}
  (\bibinfo{year}{2007}).

\bibitem{tovey24a}
\bibinfo{author}{Tovey, S.}, \bibinfo{author}{Lohrmann, C.} \&
  \bibinfo{author}{Holm, C.}
\newblock \bibinfo{title}{Emergence of chemotactic strategies with multi-agent
  reinforcement learning}.
\newblock \emph{\bibinfo{journal}{Machine Learning: Science and Technology}}
  \textbf{\bibinfo{volume}{5}}, \bibinfo{pages}{035054} (\bibinfo{year}{2024}).

\bibitem{tovey25c}
\bibinfo{author}{Tovey, S.} \emph{et~al.}
\newblock \bibinfo{title}{{{SwarmRL}}: Building the future of smart active
  systems}.
\newblock \emph{\bibinfo{journal}{The European Physical Journal E}}
  \textbf{\bibinfo{volume}{48}}, \bibinfo{pages}{16} (\bibinfo{year}{2025}).

\bibitem{freund14a}
\bibinfo{author}{Freund, J.~B.}
\newblock \bibinfo{title}{Numerical simulation of flowing blood cells}.
\newblock \emph{\bibinfo{journal}{Annual Review of Fluid Mechanics}}
  \textbf{\bibinfo{volume}{46}}, \bibinfo{pages}{67--95}
  (\bibinfo{year}{2014}).

\bibitem{marsden15a}
\bibinfo{author}{Marsden, A.~L.} \& \bibinfo{author}{{Esmaily-Moghadam}, M.}
\newblock \bibinfo{title}{Multiscale modeling of cardiovascular flows for
  clinical decision support}.
\newblock \emph{\bibinfo{journal}{Applied Mechanics Reviews}}
  \textbf{\bibinfo{volume}{67}} (\bibinfo{year}{2015}).

\bibitem{britannica26a}
\bibinfo{author}{Britannica, E.}
\newblock \bibinfo{title}{Blood vessel \textbar{} definition, anatomy,
  function, \& types \textbar{} britannica}.
\newblock
  \bibinfo{howpublished}{https://www.britannica.com/science/blood-vessel}
  (\bibinfo{year}{2026}).

\bibitem{betts13a}
\bibinfo{author}{Betts, J.~G.} \emph{et~al.}
\newblock \bibinfo{title}{Ch. 1 introduction - anatomy and physiology
  \textbar{} {{OpenStax}}}.
\newblock
  \bibinfo{howpublished}{https://assets.openstax.org/oscms-prodcms/media/documents/anatomy-and-physiology-2e\_-\_WEB.pdf}
  (\bibinfo{year}{2013}).

\bibitem{bird02a}
\bibinfo{author}{Bird, R.}, \bibinfo{author}{Stewart, W.} \&
  \bibinfo{author}{Lightfoot, E.}
\newblock \emph{\bibinfo{title}{Transport Phenomena}}
  (\bibinfo{publisher}{{John Wiley and Sons}}, \bibinfo{address}{New York},
  \bibinfo{year}{2002}), \bibinfo{edition}{2} edn.

\bibitem{maaten08a}
\bibinfo{author}{van~der Maaten, L.} \& \bibinfo{author}{Hinton, G.}
\newblock \bibinfo{title}{Visualizing {{Data}} using t-{{SNE}}}.
\newblock \emph{\bibinfo{journal}{Journal of Machine Learning Research}}
  \textbf{\bibinfo{volume}{9}}, \bibinfo{pages}{2579--2605}
  (\bibinfo{year}{2008}).

\bibitem{pedregosa11a}
\bibinfo{author}{Pedregosa, F.} \emph{et~al.}
\newblock \bibinfo{title}{Scikit-learn: Machine learning in python}.
\newblock \emph{\bibinfo{journal}{Journal of Machine Learning Research}}
  \textbf{\bibinfo{volume}{12}}, \bibinfo{pages}{2825--2830}
  (\bibinfo{year}{2011}).

\bibitem{lloyd82a}
\bibinfo{author}{Lloyd, S.}
\newblock \bibinfo{title}{Least squares quantization in {{PCM}}}.
\newblock \emph{\bibinfo{journal}{IEEE Transactions on Information Theory}}
  \textbf{\bibinfo{volume}{28}}, \bibinfo{pages}{129--137}
  (\bibinfo{year}{1982}).

\bibitem{rousseeuw87a}
\bibinfo{author}{Rousseeuw, P.~J.}
\newblock \bibinfo{title}{Silhouettes: {{A}} graphical aid to the
  interpretation and validation of cluster analysis}.
\newblock \emph{\bibinfo{journal}{Journal of Computational and Applied
  Mathematics}} \textbf{\bibinfo{volume}{20}}, \bibinfo{pages}{53--65}
  (\bibinfo{year}{1987}).

\bibitem{phillips12a}
\bibinfo{author}{Phillips, R.}, \bibinfo{author}{Kondev, J.},
  \bibinfo{author}{Theriot, J.} \& \bibinfo{author}{Garcia, H.}
\newblock \bibinfo{title}{What and where: Construction plans for cells and
  organisms}.
\newblock In \emph{\bibinfo{booktitle}{Physical Biology of the Cell}}, 2,
  \bibinfo{pages}{68} (\bibinfo{publisher}{Garland Science},
  \bibinfo{address}{New York}, \bibinfo{year}{2012}), \bibinfo{edition}{2} edn.

\bibitem{kannojiya21a}
\bibinfo{author}{Kannojiya, V.}, \bibinfo{author}{Das, A.~K.} \&
  \bibinfo{author}{Das, P.~K.}
\newblock \bibinfo{title}{Simulation of blood as fluid: A review from
  rheological aspects}.
\newblock \emph{\bibinfo{journal}{IEEE Reviews in Biomedical Engineering}}
  \textbf{\bibinfo{volume}{14}}, \bibinfo{pages}{327--341}
  (\bibinfo{year}{2021}).

\bibitem{ascolese19a}
\bibinfo{author}{Ascolese, M.}, \bibinfo{author}{Farina, A.} \&
  \bibinfo{author}{Fasano, A.}
\newblock \bibinfo{title}{The {{F\aa hr\ae us-Lindqvist}} effect in small blood
  vessels: How does it help the heart?}
\newblock \emph{\bibinfo{journal}{Journal of Biological Physics}}
  \textbf{\bibinfo{volume}{45}}, \bibinfo{pages}{379--394}
  (\bibinfo{year}{2019}).

\bibitem{bollinger74a}
\bibinfo{author}{Bollinger, A.}, \bibinfo{author}{Butti, P.},
  \bibinfo{author}{Barras, J.~P.}, \bibinfo{author}{Trachsler, H.} \&
  \bibinfo{author}{Siegenthaler, W.}
\newblock \bibinfo{title}{Red blood cell velocity in nailfold capillaries of
  man measured by a television microscopy technique}.
\newblock \emph{\bibinfo{journal}{Microvascular Research}}
  \textbf{\bibinfo{volume}{7}}, \bibinfo{pages}{61--72} (\bibinfo{year}{1974}).

\bibitem{hudetz97a}
\bibinfo{author}{Hudetz, A.~G.}
\newblock \bibinfo{title}{Blood flow in the cerebral capillary network: A
  review emphasizing observations with intravital microscopy}.
\newblock \emph{\bibinfo{journal}{Microcirculation}}
  \textbf{\bibinfo{volume}{4}}, \bibinfo{pages}{233--252}
  (\bibinfo{year}{1997}).

\bibitem{jarolimova25a-pre}
\bibinfo{author}{Jarol{\'i}mov{\'a}, A.} \emph{et~al.}
\newblock \bibinfo{title}{In vivo evidence of blood flow slippage: Failure of
  the no-slip boundary condition assumption} (\bibinfo{year}{2025}).
\newblock \eprint{2510.18107}.

\bibitem{koutsiaris03a}
\bibinfo{author}{Koutsiaris, A.~G.} \& \bibinfo{author}{Pogiatzi, A.}
\newblock \bibinfo{title}{Velocity pulse measurements in the mesenteric
  arterioles of rabbits}.
\newblock \emph{\bibinfo{journal}{Physiological Measurement}}
  \textbf{\bibinfo{volume}{25}}, \bibinfo{pages}{15} (\bibinfo{year}{2003}).

\bibitem{nader19a}
\bibinfo{author}{Nader, E.} \emph{et~al.}
\newblock \bibinfo{title}{Blood rheology: Key parameters, impact on blood flow,
  role in sickle cell disease and effects of exercise}.
\newblock \emph{\bibinfo{journal}{Frontiers in Physiology}}
  \textbf{\bibinfo{volume}{10}} (\bibinfo{year}{2019}).

\bibitem{JAX}
\bibinfo{author}{Bradbury, J.} \emph{et~al.}
\newblock \bibinfo{title}{{JAX}: composable transformations of
  {P}ython+{N}um{P}y programs} (\bibinfo{year}{2018}).
\newblock \urlprefix\url{http://github.com/google/jax}.

\bibitem{flax}
\bibinfo{author}{Heek, J.} \emph{et~al.}
\newblock \bibinfo{title}{{F}lax: A neural network library and ecosystem for
  {JAX}} (\bibinfo{year}{2023}).
\newblock \urlprefix\url{http://github.com/google/flax}.

\bibitem{weik19a}
\bibinfo{author}{Weik, F.} \emph{et~al.}
\newblock \bibinfo{title}{{{ESPResSo}} 4.0 -- an extensible software package
  for simulating soft matter systems}.
\newblock \emph{\bibinfo{journal}{The European Physical Journal Special
  Topics}} \textbf{\bibinfo{volume}{227}}, \bibinfo{pages}{1789--1816}
  (\bibinfo{year}{2019}).

\bibitem{bauer21b}
\bibinfo{author}{Bauer, M.} \emph{et~al.}
\newblock \bibinfo{title}{{{waLBerla}}: {{A}} block-structured high-performance
  framework for multiphysics simulations}.
\newblock \emph{\bibinfo{journal}{Computers \& Mathematics with Applications}}
  \textbf{\bibinfo{volume}{81}}, \bibinfo{pages}{478--501}
  (\bibinfo{year}{2021}).

\bibitem{liu97a}
\bibinfo{author}{Liu, M.}, \bibinfo{author}{Nicholson, J.~K.},
  \bibinfo{author}{Parkinson, J.~A.} \& \bibinfo{author}{Lindon, J.~C.}
\newblock \bibinfo{title}{Measurement of biomolecular diffusion coefficients in
  blood plasma using two-dimensional{\textsuperscript{1}}
  {{H}}-{\textsuperscript{1}} {{H}} diffusion-edited total-correlation {{NMR}}
  spectroscopy}.
\newblock \emph{\bibinfo{journal}{Analytical Chemistry}}
  \textbf{\bibinfo{volume}{69}}, \bibinfo{pages}{1504--1509}
  (\bibinfo{year}{1997}).

\bibitem{virtanen20b}
\bibinfo{author}{Virtanen, P.} \emph{et~al.}
\newblock \bibinfo{title}{{{SciPy}} 1.0: Fundamental algorithms for scientific
  computing in {{Python}}}.
\newblock \emph{\bibinfo{journal}{Nature Methods}}
  \textbf{\bibinfo{volume}{17}}, \bibinfo{pages}{261--272}
  (\bibinfo{year}{2020}).

\bibitem{weeks71b}
\bibinfo{author}{Weeks, J.~D.}, \bibinfo{author}{Chandler, D.} \&
  \bibinfo{author}{Andersen, H.~C.}
\newblock \bibinfo{title}{Role of repulsive forces in determining the
  equilibrium structure of simple liquids}.
\newblock \emph{\bibinfo{journal}{The Journal of Chemical Physics}}
  \textbf{\bibinfo{volume}{54}}, \bibinfo{pages}{5237--5247}
  (\bibinfo{year}{1971}).

\bibitem{barto83a}
\bibinfo{author}{Barto, A.~G.}, \bibinfo{author}{Sutton, R.~S.} \&
  \bibinfo{author}{Anderson, C.~W.}
\newblock \bibinfo{title}{Neuronlike adaptive elements that can solve difficult
  learning control problems}.
\newblock \emph{\bibinfo{journal}{IEEE Transactions on Systems, Man, and
  Cybernetics}} \textbf{\bibinfo{volume}{SMC-13}}, \bibinfo{pages}{834--846}
  (\bibinfo{year}{1983}).

\bibitem{grondman12a}
\bibinfo{author}{Grondman, I.}, \bibinfo{author}{Busoniu, L.},
  \bibinfo{author}{Lopes, G. A.~D.} \& \bibinfo{author}{Babuska, R.}
\newblock \bibinfo{title}{A survey of actor-critic reinforcement learning:
  Standard and natural policy gradients}.
\newblock \emph{\bibinfo{journal}{IEEE Transactions on Systems, Man, and
  Cybernetics, Part C (Applications and Reviews)}}
  \textbf{\bibinfo{volume}{42}}, \bibinfo{pages}{1291--1307}
  (\bibinfo{year}{2012}).

\bibitem{schulman17a-pre}
\bibinfo{author}{Schulman, J.}, \bibinfo{author}{Wolski, F.},
  \bibinfo{author}{Dhariwal, P.}, \bibinfo{author}{Radford, A.} \&
  \bibinfo{author}{Klimov, O.}
\newblock \bibinfo{title}{Proximal policy optimization algorithms}
  (\bibinfo{year}{2017}).
\newblock \eprint{1707.06347}.

\bibitem{schulman18a-pre}
\bibinfo{author}{Schulman, J.}, \bibinfo{author}{Moritz, P.},
  \bibinfo{author}{Levine, S.}, \bibinfo{author}{Jordan, M.~I.} \&
  \bibinfo{author}{Abbeel, P.}
\newblock \bibinfo{title}{High-dimensional continuous control using generalized
  advantage estimation} (\bibinfo{year}{2018}).
\newblock \eprint{1506.02438}.

\bibitem{huber64a}
\bibinfo{author}{Huber, P.~J.}
\newblock \bibinfo{title}{Robust estimation of a location parameter}.
\newblock \emph{\bibinfo{journal}{The Annals of Mathematical Statistics}}
  \textbf{\bibinfo{volume}{35}}, \bibinfo{pages}{73--101}
  (\bibinfo{year}{1964}).

\bibitem{gronauer22a}
\bibinfo{author}{Gronauer, S.} \& \bibinfo{author}{Diepold, K.}
\newblock \bibinfo{title}{Multi-agent deep reinforcement learning: A survey}.
\newblock \emph{\bibinfo{journal}{Artificial Intelligence Review}}
  \textbf{\bibinfo{volume}{55}}, \bibinfo{pages}{895--943}
  (\bibinfo{year}{2022}).

\bibitem{oliehoek16a}
\bibinfo{author}{Oliehoek, F.~A.} \& \bibinfo{author}{Amato, C.}
\newblock \emph{\bibinfo{title}{A Concise Introduction to Decentralized
  {{POMDPs}}}}.
\newblock {{SpringerBriefs}} in {{Intelligent Systems}}
  (\bibinfo{publisher}{Springer International Publishing},
  \bibinfo{address}{Cham}, \bibinfo{year}{2016}).

\bibitem{kingma17a}
\bibinfo{author}{Kingma, D.~P.} \& \bibinfo{author}{Ba, J.}
\newblock \bibinfo{title}{Adam: {{A Method}} for {{Stochastic Optimization}}}.
\newblock \emph{\bibinfo{journal}{arXiv}}  (\bibinfo{year}{2017}).
\newblock \eprint{1412.6980}.

\bibitem{murray92a}
\bibinfo{author}{Murray, A.~G.} \& \bibinfo{author}{Jackson, G.~A.}
\newblock \bibinfo{title}{Viral dynamics: A model of the effects of size,
  shape, motion and abundance of single-celled planktonic organisms and other
  particles}.
\newblock \emph{\bibinfo{journal}{Marine Ecology Progress Series}}
  \textbf{\bibinfo{volume}{89}}, \bibinfo{pages}{103--116}
  (\bibinfo{year}{1992}).

\end{thebibliography}
\end{document}